%% file: main.tex
\documentclass{article}
\usepackage[utf8]{inputenc}
\usepackage{main}
\usepackage{microtype}
\usepackage{graphicx}
\usepackage{subfig}
\usepackage{times}
\usepackage{latexsym}
\usepackage{amsmath}
\usepackage{float}
\usepackage{footnote}
\usepackage{enumitem}
\usepackage{bm}
\usepackage{arydshln}
\usepackage{booktabs}
\usepackage{multicol}
\usepackage{multirow}
\usepackage{color}
\usepackage{xcolor}     
\usepackage{colortbl}
\usepackage{bbding}
\usepackage{makecell}
\usepackage{mathtools}
\usepackage{imakeidx}
\usepackage{longtable}
\usepackage{wrapfig}
\makeindex
\usepackage{arydshln}
\usepackage{lipsum}
\usepackage{natbib}
\usepackage[toc]{multitoc}
\usepackage[edges]{forest}
\usepackage[normalem]{ulem}
\definecolor{mydarkblue}{rgb}{0,0.08,0.45}
\usepackage[colorlinks=true,linkcolor=mydarkblue,citecolor=mydarkblue,filecolor=mydarkblue,urlcolor=mydarkblue]{hyperref}
\usepackage{caption}
\usepackage{CJKutf8}
\usepackage{awesomebox} 
\usepackage{bbding}
\usepackage[most]{tcolorbox}

\input{pre}

\usepackage[tikz]{bclogo}
\usepackage[framemethod=tikz]{mdframed}
\definecolor{bgblue}{RGB}{245,243,253}
\definecolor{ttblue}{RGB}{91,194,224}

\mdfdefinestyle{mystyle}{%
  rightline=true,
  innerleftmargin=10,
  innerrightmargin=10,
  outerlinewidth=3pt,
  topline=false,
  rightline=true,
  bottomline=false,
  skipabove=\topsep,
  skipbelow=\topsep
}

\newtcolorbox{myboxi}[1][]{
  breakable,
  title=#1,
  colback=red!5,
  colbacktitle=red!5,
  coltitle=black,
  fonttitle=\bfseries,
  bottomrule=0pt,
  toprule=0pt,
  leftrule=2pt,
  rightrule=2pt,
  titlerule=0pt,
  arc=0pt,
  outer arc=0pt,
  colframe=red,
}

\newtcolorbox{myboxnote}[1][]{
  breakable,
  title=#1,
  colback=orange!0,
  colbacktitle=orange!0,
  coltitle=black,
  fonttitle=\bfseries,
  bottomrule=0pt,
  toprule=0pt,
  leftrule=2pt,
  rightrule=2pt,
  titlerule=0pt,
  arc=0pt,
  outer arc=0pt,
  colframe=orange,
}

\newtcolorbox{myboxii}[1][]{
  breakable,
  freelance,
  title=#1,
  colback=white,
  colbacktitle=white,
  coltitle=black,
  fonttitle=\bfseries,
  bottomrule=0pt,
  boxrule=0pt,
  colframe=white,
  overlay unbroken and first={
  \draw[red!75!black,line width=3pt]
    ([xshift=5pt]frame.north west) -- 
    (frame.north west) -- 
    (frame.south west);
  \draw[red!75!black,line width=3pt]
    ([xshift=-5pt]frame.north east) -- 
    (frame.north east) -- 
    (frame.south east);
  },
  overlay unbroken app={
  \draw[red!75!black,line width=3pt,line cap=rect]
    (frame.south west) -- 
    ([xshift=5pt]frame.south west);
  \draw[red!75!black,line width=3pt,line cap=rect]
    (frame.south east) -- 
    ([xshift=-5pt]frame.south east);
  },
  overlay middle and last={
  \draw[red!75!black,line width=3pt]
    (frame.north west) -- 
    (frame.south west);
  \draw[red!75!black,line width=3pt]
    (frame.north east) -- 
    (frame.south east);
  },
  overlay last app={
  \draw[red!75!black,line width=3pt,line cap=rect]
    (frame.south west) --
    ([xshift=5pt]frame.south west);
  \draw[red!75!black,line width=3pt,line cap=rect]
    (frame.south east) --
    ([xshift=-5pt]frame.south east);
  },
}

\usepackage{fancyhdr} 
\usepackage{blindtext} 

\DeclareCaptionFont{black}{\color{black}}

\definecolor{myblue}{rgb}{0.9, 0.1, 0.94}
\definecolor{mygreen}{rgb}{0.64, 0.56, 0.88}
\definecolor{myyellow}{rgb}{0.68, 0.6, 0.1}
\definecolor{fancygreen}{rgb}{0.33, 0.68, 0.20}
\definecolor{salmon}{rgb}{0.94, 0.52, 0.49}
\definecolor{tablegreen}{rgb}{0.82, 0.94, 0.75}
\definecolor{tableblue}{rgb}{0.81, 0.90, 0.94}
\definecolor{tablered}{rgb}{0.97, 0.85, 0.85}
\definecolor{tableorange}{rgb}{0.96, 0.85, 0.81}

\newenvironment{itemize*}%
 {\leftmargini=10pt\begin{itemize}%
  \setlength{\itemsep}{0pt}%
  \setlength{\parskip}{0pt}%
  }%
 {\end{itemize}}
\newenvironment{enumerate*}%
 {\begin{enumerate}%
  \setlength{\itemsep}{0pt}%
  \setlength{\parskip}{0pt}}%
 {\end{enumerate}}

\usepackage{xcolor}
\usepackage{listings}

\newcommand\JSONnumbervaluestyle{\color{blue}}
\newcommand\JSONstringvaluestyle{\color{red}}

\newif\ifcolonfoundonthisline

\makeatletter

\lstdefinestyle{json}
{
  showstringspaces    = false,
  keywords            = {false,true},
  alsoletter          = 0123456789.,
  morestring          = [s]{"}{"},
  stringstyle         = \ifcolonfoundonthisline\JSONstringvaluestyle\fi,
  MoreSelectCharTable =%
    \lst@DefSaveDef{`:}\colon@json{\processColon@json},
  basicstyle          = \ttfamily,
  keywordstyle        = \ttfamily\bfseries,
}

\newcommand\processColon@json{%
  \colon@json%
  \ifnum\lst@mode=\lst@Pmode%
    \global\colonfoundonthislinetrue%
  \fi
}

\lst@AddToHook{Output}{%
  \ifcolonfoundonthisline%
    \ifnum\lst@mode=\lst@Pmode%
      \def\lst@thestyle{\JSONnumbervaluestyle}%
    \fi
  \fi
  \lsthk@DetectKeywords%
}

\lst@AddToHook{EOL}%
  {\global\colonfoundonthislinefalse}

\makeatother

\usepackage{etoolbox}
\usepackage{natbib}
\usepackage{url}
\newcounter{bibcount}
\makeatletter
\patchcmd{\@lbibitem}{\item[}{\item[\hfil\stepcounter{bibcount}{[\thebibcount]}}{}{}
\renewcommand\NAT@bibsetup%
  [1]{\setlength{\leftmargin}{\bibhang}\setlength{\itemindent}{-\parindent}%
      \setlength{\itemsep}{\bibsep}\setlength{\parsep}{\z@}}
\makeatother

\begin{document}

\title{\ourbench: Benchmarking Agent Systems for Demand-Driven Dataset Discovery}

\author{
\textbf{Keyu Li}$^{1,2}$\thanks{Equal contribution} \quad
\textbf{Mohan Jiang}$^{1,2*}$ \quad
\textbf{Dayuan Fu}$^{2,3}$ \quad
\textbf{Yunze Wu}$^{1,2,3}$ \quad 
\textbf{Xiangkun Hu}$^{2,3}$ \quad \\
\textbf{Dequan Wang}$^{1,2}$\thanks{Corresponding author} \quad
\textbf{Pengfei Liu}$^{1,2,3\dagger}$\\
\textsuperscript{1}Shanghai Jiao Tong University\quad
\textsuperscript{2}SII\quad
\textsuperscript{3}GAIR \\
\texttt{\{chlorophyll, mhjiang0408, dequanwang, pengfei\}@sjtu.edu.cn} \\\\
    \github \href{https://github.com/GAIR-NLP/DatasetResearch}{\textbf{GAIR-NLP/DatasetResearch}} ~ ~ ~ \huggingface \href{https://huggingface.co/datasets/GAIR/DatasetResearch}{\textbf{GAIR/DatasetResearch}} 
}

\maketitle
\thispagestyle{fancy}
\fancyhead{}
\lhead{\includegraphics[height=0.90cm]{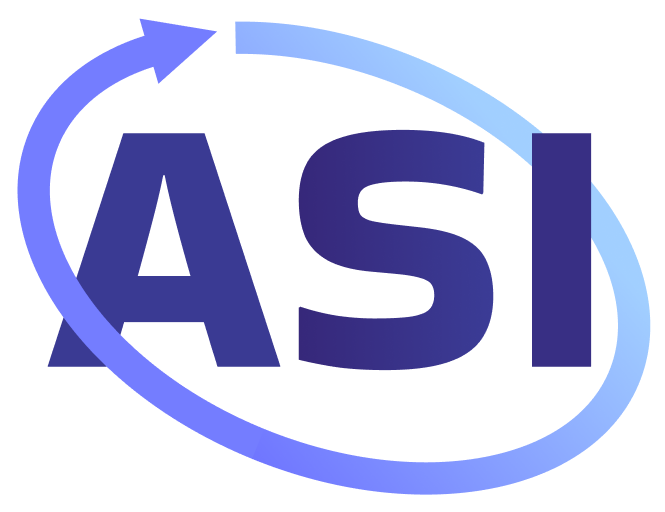}}
\rhead{%
  \raisebox{-0.1cm}{\includegraphics[height=0.7cm]{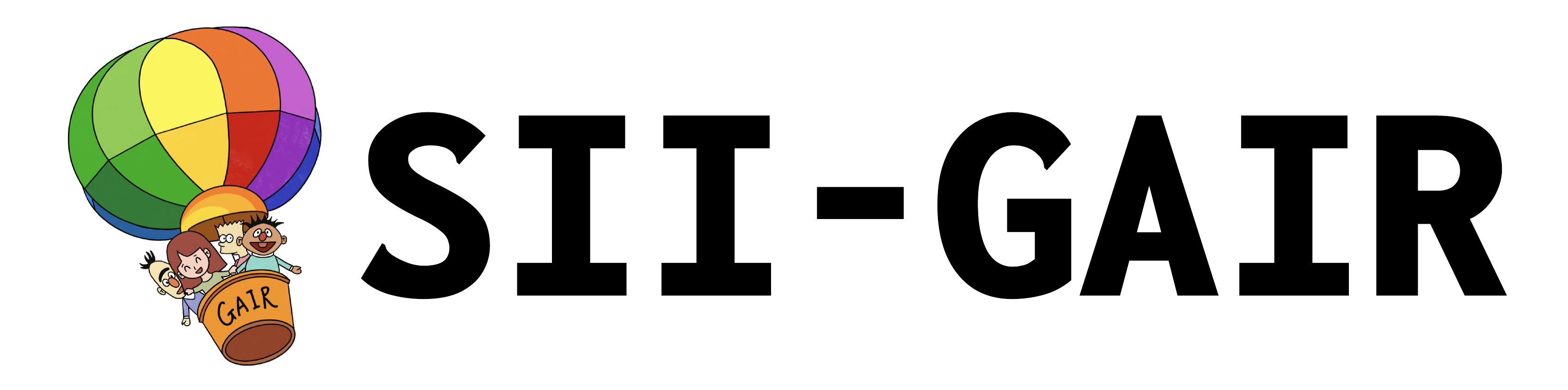}}%
}
\renewcommand{\headrulewidth}{0pt}
\setlength{\headsep}{3mm} 

\vspace{-1.0em}

\input{sec/abstract}

\newpage

\pagestyle{fancy}
\lhead{\rightmark}
\renewcommand{\headrulewidth}{0.7pt}
\setlength{\headsep}{5mm}


\clearpage

\input{sec/intro}

\input{sec/related}
\input{sec/method}

\input{sec/experiment}

\input{sec/analysis}
\input{sec/conclusion}

\newpage
\bibliographystyle{unsrtnat}
\bibliography{ref}

\newpage

\input{sec/appendix}

\end{document}

%% file: pre.tex
\definecolor{gred}{RGB}{250, 210, 207}

\definecolor{coolblue1}{rgb}{0.91, 0.94, 0.98} 
\definecolor{coolblue2}{rgb}{0.76, 0.85, 0.94} 
\definecolor{coolblue3}{rgb}{0.54, 0.72, 0.87} 
\definecolor{coolblue4}{rgb}{1, 1, 1} 

\newcommand{\ourbench}{\textsc{DatasetResearch}\xspace}

\newcommand{\ourbenchmini}{\texttt{DatasetResearch-pro}\xspace}

\newcommand{\ourbenchnum}{208\xspace}

\newcommand{\llama}{LLaMA-3.1-8B\xspace}

\newcommand{\ourbaseline}{DataResearcher\xspace}

\newcommand{\github}{\raisebox{-1.5pt}{\includegraphics[height=1.05em]{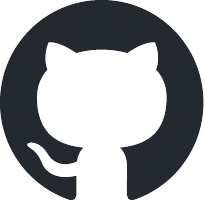}}\xspace}
\newcommand{\huggingface}{\raisebox{-1.5pt}{\includegraphics[height=1.05em]{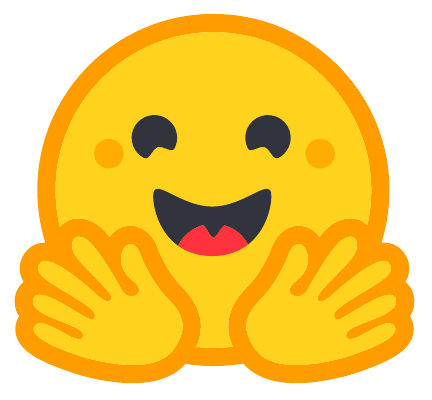}}\xspace}

\usepackage{cleveref}

%% file: sec/abstract.tex
\begin{abstract}

    The rapid advancement of large language models has fundamentally shifted the bottleneck in AI development from computational power to data availability—with countless valuable datasets remaining hidden across specialized repositories, research appendices, and domain platforms. As reasoning capabilities and deep research methodologies continue to evolve, a critical question emerges: can AI agents transcend conventional search to systematically discover any dataset that meets specific user requirements, enabling truly autonomous demand-driven data curation?
    We introduce \ourbench, the first comprehensive benchmark evaluating AI agents' ability to discover and synthesize datasets from \ourbenchnum real-world demands across knowledge-intensive and reasoning-intensive tasks. Our tri-dimensional evaluation framework reveals a stark reality: even advanced deep research systems achieve only 22\% score on our challenging \ourbenchmini subset, exposing the vast gap between current capabilities and perfect dataset discovery.
    Our analysis uncovers a fundamental dichotomy—search agents excel at knowledge tasks through retrieval breadth, while synthesis agents dominate reasoning challenges via structured generation—yet both catastrophically fail on ``corner cases'' outside existing distributions. These findings establish the first rigorous baseline for dataset discovery agents and illuminate the path toward AI systems capable of finding any dataset in the digital universe.Our benchmark and comprehensive analysis provide the foundation for the next generation of self-improving AI systems and are publicly available at \url{https://github.com/GAIR-NLP/DatasetResearch}.
\end{abstract}

\input{figs/tex/teaser}

%% file: figs/tex/teaser.tex
\begin{figure*}[ht]
    \centering
    \includegraphics[width=0.9\linewidth,height=0.45\linewidth]{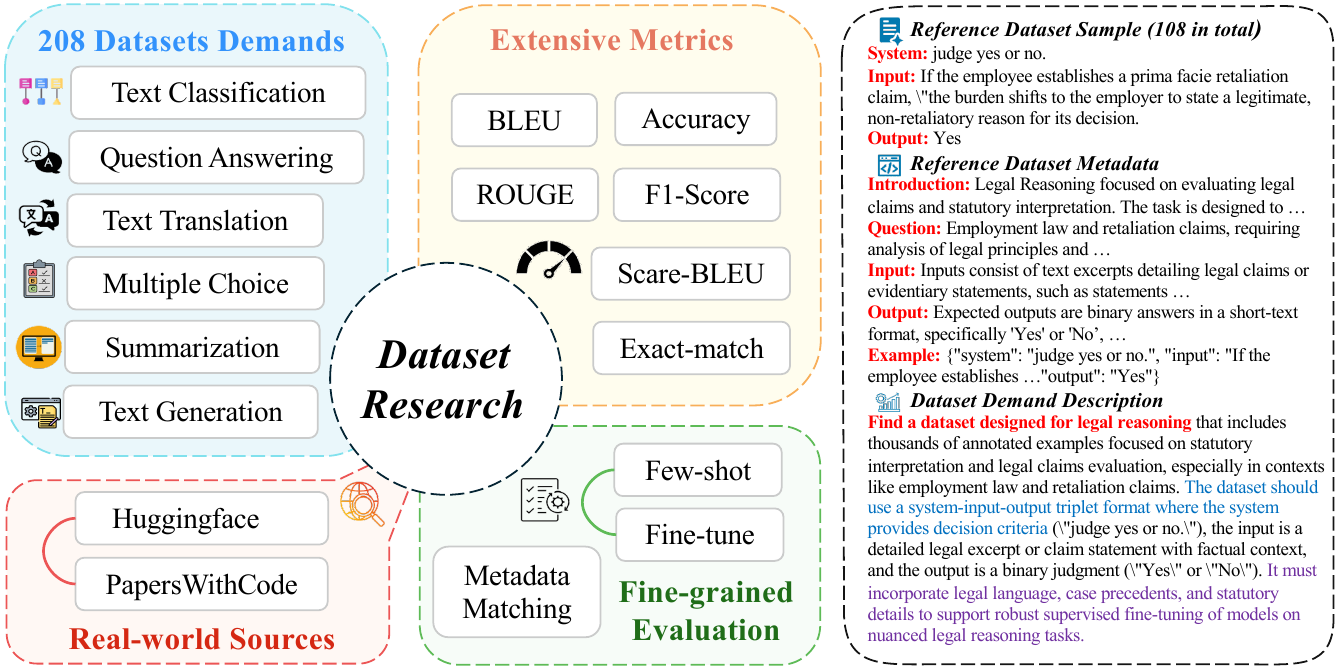}
    \caption{\textbf{Overview of \ourbench:} A Benchmark for Dataset Discovery Agents. It features two sources, over 200 dataset demands, three evaluation methodologies, and six NLP evaluation metrics.}
    \label{fig:teaser}
\end{figure*}

%% file: sec/intro.tex
\section{Introduction}

The advancement of artificial intelligence (AI)~\cite{grok4, o3, claude4, comanici2025gemini, guo2025deepseek} has increasingly positioned datasets as foundational assets for scientific discovery and technological progress. Contemporary powerful reinforcement learning-based agentic systems exhibit a strong dependence on high-quality datasets for optimal performance~\cite{ye2025limo,wang2025octothinker,sun2024llm,wang2024reinforcement,team2025kimi,mai2025agent,yang2025swe,xie2025agentsynth,wei2025browsecomp}. The identification and synthesis of appropriate datasets represents a critical bottleneck in the initiation of scientific research endeavors.
The traditional research workflow—problem identification, dataset requirement formulation, manual data search, synthesis, and model training—represents an increasingly antiquated approach in an era of rapidly evolving AI capabilities. As reasoning methodologies and deep research tools continue to advance, a fundamental question emerges: can AI agents transcend the limitations of conventional search to systematically discover any dataset that meets specific user requirements from this vast, largely untapped data universe?


While previous work has explored demand-driven dataset discovery and synthesis~\cite{viswanathan2023datafinder,walker2023prompting,gandhi2024better}, these approaches represent initial steps that do not fully harness the deep reasoning and inferential capabilities enabled by modern large language models.
The emergence of sophisticated reasoning agents and deep research methodologies demands a more rigorous evaluation of whether current systems can achieve the ambitious goal of comprehensive, demand-driven dataset discovery across the entire digital landscape.
To address this critical gap, we introduce \ourbench, the first comprehensive benchmark designed to systematically evaluate whether AI agents can make any relevant dataset discoverable based on specific user demands. As the inaugural benchmark in this domain, \ourbench exhibits several distinctive characteristics shown in \Cref{fig:teaser}:

\begin{itemize*}
\item \textbf{Comprehensive Coverage and Rich Examples:} We curated \ourbenchnum data requirements suitable for fine-tuning tasks and their corresponding datasets from Huggingface and PaperswithCode, spanning six major Natural Language Processing (NLP) tasks. For each dataset, we employed OpenAI o3~\cite{o3} to generate comprehensive metadata including task descriptions, questions, input-output specifications, and examples. To capture diverse cognitive demands, we classify all requirements into knowledge-based and reasoning-based categories shown in \Cref{tab:dataset_info}, distinguishing between needs for factual information versus complex inference. Based on this metadata, we generated corresponding query pairs for search and synthesis agents, serving as dataset demand inputs.
\item \textbf{Thorough Evaluation with Diverse Baselines:} 
We construct \ourbaseline to comprehensively evaluate the quality of data constructed by search agents, synthesis agents and deep research agents. We also designed an integrated evaluation methodology encompassing three assessment approaches: metadata evaluation, few-shot evaluation, and supervised fine-tuning. This evaluation framework quantifies the construction performance of \ourbaseline across \ourbenchnum demand tasks through a normalized score. Our evaluation encompasses two search agent baselines including GPT-4o-search-preview~\cite{gpt-4o-search-preview}, two synthesis agent settings with OpenAI o3~\cite{o3} and three deep research agents~\cite{grokdeepresearch,openaideepresearch,geminideepresearch}. Evaluation employs metadata similarity and \llama~\cite{dubey2024llama} for few-shot and instruction-tuning~\cite{ouyang2022training} followed by assessment on original reference sets.
\item \textbf{Stratified Difficulty with Reference Subsets:} Given the computational resource demands of fine-tuning and the high computational costs of deep research systems, we carefully curated 20 more challenging data construction demands from \ourbench to construct the \ourbenchmini subset. This subset construction is based on preliminary experimental results, selecting tasks that proved most challenging for GPT-4o-search-preview, thereby ensuring effective differentiation of capability boundaries across different agent systems.
\end{itemize*}

Comprehensive experiments on \ourbench demonstrate that current agent systems fall considerably short of optimal performance, with even the most advanced deep research systems achieving a maximum score of merely 22\% on our evaluation subsets. Our analysis further reveals a pronounced performance differentiation pattern: search agents leverage their robust information retrieval capabilities to excel in knowledge-based tasks, while synthesis agents capitalize on their capacity for constructing reasoning pathways to demonstrate superior performance in reasoning-based challenges. These findings not only expose the limitations of existing technologies but also highlight the tremendous potential of automated data synthesis approaches.

To conclude, we make the following contributions:


\begin{itemize*}
\item We present \ourbench, the first comprehensive benchmark for systematically evaluating agent systems' capabilities in demand-driven dataset discovery and synthesis, comprising \ourbenchnum real-world dataset requirements from HuggingFace and PaperswithCode.

\item We develop a multi-dimensional evaluation methodology encompassing metadata alignment assessment, few-shot performance evaluation, and supervised fine-tuning effectiveness across six task categories.

\item Our extensive experiments on state-of-the-art systems reveal significant limitations, with maximum scores of only 0.2 on \ourbenchmini. We identify distinct performance patterns where search agents excel in knowledge-based tasks while synthesis agents demonstrate advantages in reasoning-based challenges.

\item We provide the first systematic analysis of failure modes in automated dataset construction, revealing that all current methods struggle with corner cases outside existing data distributions, highlighting fundamental challenges in generalization.
\end{itemize*}

%% file: sec/related.tex
\section{Related Work}

\subsection{Dataset Discovery and Synthesis}
The critical importance of high-quality datasets for model training has been increasingly recognized in recent research \cite{ye2025limo, wang2025octothinker,yang2025swe,xie2025agentsynth,chen2024scienceagentbench}, underscoring the unique significance of data in the rapidly evolving AI landscape. Given a specific dataset requirement, the task of identifying appropriate datasets for scientific research has been partially explored through two primary research directions: dataset discovery and dataset synthesis.

\paragraph{Dataset Discovery} 
In the domain of dataset discovery, several approaches have been proposed to automate the retrieval process. \citet{viswanathan2023datafinder,soylu2024fine} introduced a bi-encoder retriever system that recommends datasets based on natural language research descriptions, demonstrating the potential for semantic matching between research queries and dataset characteristics. Similarly, \cite{walker2023prompting,majumder2024discoverybench,gu2024blade,yang2025impact} explored the use of conversational generative AI for data discovery, revealing that while large language models can recommend relevant datasets and provide coherent reasoning, they are prone to hallucinating non-existent datasets, which highlights the need to evaluate their ability to acquire and practically apply data based on real-world demands.

\paragraph{Dataset Synthesis} 
For dataset synthesis, recent work has focused on transforming and repurposing existing resources. \citet{gandhi2024better} introduced dataset transformation techniques to repurpose existing public datasets for target tasks, achieving substantial improvements over both synthetic data generation and few-shot prompting baselines. This approach demonstrates the value of leveraging existing data structures while adapting them to new requirements. Using a suite of agentic methods, \citet{yang2025swe} constructed a high-quality dataset and achieved a pass@1 result of 40.2\% on the SWE-bench verified~\cite{jimenez2023swe}  with Qwen-2.5-Coder-Instruct-32B~\cite{qwen2,hui2024qwen2}.

The emergence of increasingly capable agent systems presents new opportunities for dataset acquisition. Modern agents can engage in iterative reasoning while conducting dataset searches, potentially uncovering even long-tail, sparse samples \cite{zheng2025deepresearcher,singh2025code}. This evolution necessitates a comprehensive benchmark to evaluate agent systems' ability to search for or synthesize datasets based on given requirements, motivating our development of \ourbench.

\subsection{Agents with Search and Reasoning Capabilities}
The continuous evolution of AI agent capabilities stems from two complementary advances: the strengthening of foundation models \cite{grok4, o3, claude4, comanici2025gemini, guo2025deepseek,yang2025qwen3} and the development of sophisticated search, reasoning, and tool-calling abilities \cite{zheng2025deepresearcher, jin2025search, song2025r1,lu2024toolsandbox}. \ourbench is designed to simultaneously evaluate both search-enabled agents and reasoning agents in their capacity to discover and synthesize datasets.

\paragraph{Search Agent and Deep Research Agent} 
Contemporary search-enabled agents encompass both LLM-based systems with integrated search tools, such as GPT-4o Search and GPT-4o-mini Search \cite{gpt-4o-search-preview}, and more sophisticated closed-source deep research systems, including OpenAI Deep Research \cite{openaideepresearch}, Gemini Deep Research \cite{geminideepresearch}, and Grok Deep Research \cite{grokdeepresearch}. These systems demonstrate varying capabilities in information retrieval and synthesis across different domains.

\paragraph{Reasoning Agent}
For reasoning-based dataset synthesis, we evaluate leading reasoning agents including OpenAI o3 \cite{o3}, Gemini 2.5 Pro \cite{comanici2025gemini}, Claude 4 \cite{claude4}, QwQ-32B\cite{qwq32b} and Grok 4 \cite{grok4}. In \ourbench, we particularly leverage the robust reasoning capabilities of OpenAI o3 \cite{o3} to construct challenging synthetic dataset generation tasks that test the limits of current agent systems. 

Our benchmark addresses dataset construction through two complementary approaches. First, we leverage state-of-the-art language models and agent systems equipped with search functionality to discover existing datasets that match specified requirements. Second, we employ advanced reasoning models to directly generate synthetic datasets tailored to given specifications.

%% file: sec/method.tex
\input{figs/tex/benchmark}
\input{tabs/dataset_info}

\section{\ourbench}

In this section, we systematically detail the construction of the \ourbench benchmark, including its composition and the comprehensive evaluation protocol designed to assess agentic systems. \ourbench curation pipeline is shown in \Cref{fig:benchmark}, and evaluation pipeline is shown in \Cref{fig:evaluation}.

\subsection{Task Definition}

In AI developing workflows, practitioners frequently encounter the challenge of identifying and collecting appropriate datasets that align with specific training requirements. This process, which we term \textbf{data discovery}, involves both systematically searching through available data resources and synthesizing new datasets as needed to satisfy given criteria and constraints. Formally, we define the data discovery task as follows: Given a natural language \textbf{demand description} $D$ that specifies the desired characteristics of a dataset, a \textbf{\ourbaseline} for data discovery must output a \textbf{discovered dataset} $S_{d} = \{d_1, d_2, ..., d_n\}$ that optimally satisfies the specified demand $D$. 

To evaluate this task, we establish a benchmark framework based on \textbf{MetaTriplets}, where each MetaTriplet $M_i = (D_i, S_{r_i}, \text{Meta}_{r_i})$ consists of three components: (1) a \textbf{demand description} $D_i$ representing a real-world data collection requirement expressed in natural language, (2) a \textbf{reference set} $S_{r_i} = \{d_{r_1}, d_{r_2}, ..., d_{r_k}\}$ containing ground truth datasets that satisfy the demand $D_i$, and (3) a \textbf{reference metadata} $\text{Meta}_{r_i}$ providing detailed information about each dataset in $S_{r_i}$, including domain specifications, format descriptions, quality metrics, and other relevant characteristics.

The evaluation process compares the data researcher's discovered dataset $S_d$ against the reference dataset $S_{r}$ using evaluations that assess both metadata relevance by generating a \textbf{Discovered Metadata} and downstream task performance by testing discovered dataset $S_d$ on the reference dataset $S_{r}$. This triplet-based evaluation framework enables systematic assessment of \ourbaseline across diverse domains and requirements, facilitating the development of AI-driven solutions that can autonomously identify and provision datasets for AI model training—thereby establishing a self-improving ecosystem where artificial intelligence systems enhance their own data discovery and curation capabilities.
\subsection{\ourbench Curation}
\ourbench collects requirements corresponding to \ourbenchnum real-world datasets shown in \Cref{tab:dataset_info}, with 91 sourced from HuggingFace and 117 from Papers with Code. Each dataset demand in \ourbench is accompanied by three key components: a detailed \textbf{demand description}, the corresponding \textbf{reference dataset}, and comprehensive \textbf{reference metadata} associated with the reference dataset.

\paragraph{Data Collection Pipeline}
We develope a systematic collection methodology guided by three key principles: ensuring real-world authenticity of dataset demands, maintaining automated evaluation feasibility, and preserving structural clarity for agent processing. Our collection process, shown in \Cref{fig:benchmark} and exemplified through HuggingFace datasets, follows a multi-stage filtering and refinement
  approach designed to produce high-quality, evaluable demands.

\begin{itemize*}
    \item \textbf{Step 1: Initial Curation} We use the HuggingFace API to identify all ``gated'' datasets, which require manual approval for access. We select these as our starting point to mitigate data leakage, as search agents cannot automatically download and process these datasets even if they are identified.

    \item \textbf{Step 2: Task and Modality Filtering} We filter this collection to retain only text-modality datasets whose annotated task fell within one of six categories: question-answering, text-summarization, text-classification, text-generation, multiple-choice, or language-translation. This step ensures the feasibility of an automated evaluation pipeline. We further exclude tasks where baseline model performance had already reached near-saturation levels, as these provide insufficient discriminative capacity for meaningful evaluation. This comprehensive filtering process yields 422 datasets.

    \item \textbf{Step 3: Documentation Quality Check} We further filter for datasets that contain comprehensive and informative README files, which serve as crucial references for generating reference metadata and demand descriptions. For demands from Papers with Code, we utilize the abstracts of corresponding papers and dataset samples instead of README files. This step results in 261 candidates.

    \item \textbf{Step 4: Fine-Tuning Suitability} We select datasets amenable to fine-tuning, excluding those designed purely for pre-training or lacking clear label columns. This leaves 104 suitable datasets.

    \item \textbf{Step 5: Automated Reformatting} For each of the 104 datasets, we prompt the OpenAI o3 model to propose a fine-tuning format template by analyzing the README and data samples (e.g., combining specific columns into input and output fields, adding a task-specific instruction).

    \item \textbf{Step 6: Human Verification} We manually review and refine these suggestions, adding instructions where necessary and removing a few datasets deemed unsuitable for fine-tuning. This process finalizes a set of 91 high-quality datasets from HuggingFace.

    \item \textbf{Step 7: Demand Description and Metadata Generation} For each of the 91 datasets, we use the o3 model to generate a comprehensive metadata profile, including an introduction, domain, input schema, output schema, and sample count. Based on this metadata, we then prompt OpenAI o3 to generate natural language demand descriptions that serve as inputs for our \textbf{\ourbaseline}.
\end{itemize*}

These 91 demands, combined with the 117 generated via a similar process from Papers with Code, form the \ourbenchnum tasks in \ourbench. From this pool, we curate a specialized subset of 20 particularly challenging tasks to create \ourbenchmini. This subset is constructed to probe the limits of current agents by selecting the 20 tasks where GPT-4o-search-preview achieved the lowest scores in the fine-tuning setting. On this highly difficult subset, we expand our evaluation to include the most advanced deep research agents. Detailed prompts used for metadata and demand generation are available in the Appendix.

\paragraph{Categorizing Knowledge-Based and Reasoning-Based Tasks}
To distinguish between tasks where reference dataset needs to be more oriented toward factual knowledge learning versus those requiring reasoning-based logical learning, we categorize these requirements into two types. Based on dataset characteristics and corresponding specific requirement descriptions, we manually identify and annotated 51 knowledge-based tasks and 157 reasoning-based tasks. \textbf{Knowledge-based} tasks require data coverage of extensive factual information, structured knowledge, and predefined classification systems, emphasizing the breadth and accuracy of the constructed data. In contrast, \textbf{reasoning-based} tasks require data that can guide the construction of reasoning pathways and logical relationships from input to output, emphasizing the achievement of cross-domain problem-solving capabilities and cognitive generalization through learning reasoning patterns rather than relying on precise coverage of domain knowledge in reference dataset.

\input{figs/tex/evaluation}

\subsection{Evaluation Methdology}

We employ a comprehensive evaluation methodology shown in \Cref{fig:evaluation} that assesses dataset quality from both intrinsic metadata characteristics and extrinsic performance on downstream tasks.

\paragraph{Metadata-Based Evaluation} To assess the primary data quality of discovered datasets, we score the semantic alignment between reference metadata and discovered metadata. Using OpenAI o3 as a judge~\cite{zheng2023judging}, we assign a score from 0 to 10 for each metadata dimension, including introduction, task, question, input, output, and example. Detail description is shown in \Cref{sec:metadata}. The final metadata score is the average of these dimensional scores. Critically, because the OpenAI o3 is used to generate both the reference and the discovered metadata, using it for evaluation systematically mitigates potential scoring biases.

\paragraph{Downstream Task Performance Evaluation} In order to assess the practical gains of discovered datasets on real-world tasks, we assess the practical utility of the datasets via test-time and training-based methods. For the six task categories in \ourbench, we design and employ six corresponding metrics. We evaluate performance across three settings:

\begin{itemize*}
\item \textbf{Zero-shot Baseline:} We directly evaluate \llama on the reference set without any fine-tuning or in-context examples to establish a performance floor.
\item \textbf{Few-shot Learning:} We provide 1, 3, and 5 examples from the discovered or synthesized datasets as in-context examples~\cite{parnami2022learningexamplessummaryapproaches} for \llama and evaluate on full reference dataset.
\item \textbf{Fine-tuning:} We fine-tune \llama on the discovered or synthesized datasets with fixed hyperparameters and then evaluate its zero-shot performance on the reference set.
\end{itemize*}

To ensure fair comparisons across tasks with different evaluation metrics, we implement a standardized normalization procedure for all performance scores. Given the heterogeneous nature of evaluation metrics across our six task categories—ranging from BLEU scores for translation tasks to accuracy for classification—direct comparison would be misleading without proper normalization.

The performance $S_{\text{ref}}$ of \llama model fine-tuned directly on the reference set serves as the upper bound for score normalization, representing the theoretical maximum performance achievable with perfect data for each specific task. For each evaluation setting (few-shot or fine-tuned), the score $S_{\text{eval}}$ of a discovered dataset is normalized against this upper bound. The final normalized score is calculated as:

$$\text{Normalized Score}= S_{\text{eval}}/S_{\text{ref}}$$
This formula positions the agent's performance on a scale from 0 to 1, or higher if the discovered dataset is superior to the reference set, enabling a fair and direct comparison of agent capabilities across all tasks in \ourbench.

\section{\ourbaseline}
\input{figs/tex/baseline}
To evaluate AI systems' capabilities in generating data for AI training, we construct three distinct baseline Data Researchers shown in \Cref{fig:baseline} for \ourbench: search agents, synthesis agents and deep research agents. These agents represent different paradigmatic approaches to data construction and are implemented as follows:

\begin{itemize*}
\item \textbf{Search Agents} are driven by prompts containing the natural language demand description. The search agent's task is to query the HuggingFace Hub and return the top five most relevant, publicly accessible dataset repository identifiers. We then programmatically iterate through this ranked list, selecting the first valid and downloadable dataset. This approach naturally prevents the agent from finding the original reference set, as all source datasets in our benchmark are gated.
\item \textbf{Synthesis Agents} are prompted with the natural language demand description. We then instruct the OpenAI o3 to generate a dataset of 500 sample pairs matching the demand's specifications. We establish two settings: with one reference dataset sample (w/ ref) and without reference dataset sample (w/o ref). In the former, we use the pre-existing sample from the reference set as a reference when generating. In the latter, we only use the demand description and do not include a sample from the reference set as a reference.
\item \textbf{Deep Research Agents} are provided with natural language demand descriptions and utilize deep research tools offered by OpenAI, Grok, and Gemini to employ reasoning capabilities for understanding requirements. They conduct an entire web-wide deep analysis of data sources to assess their alignment with the given demand, ultimately outputting an existing dataset deemed most suitable. Unfortunately, these deep research tools are not currently accessible via API calls, necessitating a human-in-the-loop approach to complete the data discovery process.
\end{itemize*}

After obtaining the preliminary discovered dataset, we leverage OpenAI o3 to automatically parse and convert all samples into a standardized fine-tuning format with complete input and output pairs, ensuring compatibility with downstream training procedures and yielding the final discovered dataset ready for evaluation.

For \ourbench, we conduct comprehensive evaluations on search-based APIs including gpt-4o-search-preview and gpt-4o-mini-search-preview~\cite{gpt-4o-search-preview}, which provide real-time web search capabilities, as well as the synthesis capabilities of the advanced reasoning-based model OpenAI o3, known for its sophisticated analytical and generation abilities. For the more challenging \ourbenchmini subset, we assess the performance of state-of-the-art closed-source deep search agents that represent the current frontier in AI-powered research capabilities, including OpenAI Deep Research~\cite{openaideepresearch}, Gemini Deep Research~\cite{geminideepresearch}, and Grok Deep Research~\cite{grokdeepresearch}.

To ensure a fair comparison, the Data Researchers have the following experimental settings:
\begin{itemize*}
\item Search agents evaluated on \ourbench are restricted to searching public datasets on HuggingFace to facilitate automated data formatting and processing. For \ourbenchmini, the search scope is expanded to the entire web, with results manually curated for relevance and accessibility.
\item For all datasets returned by search agents, we use the OpenAI o3 to parse the data into a standardized fine-tuning format and to generate their corresponding metadata, mirroring the format of our reference set. If a discovered dataset contained more than 1000 samples, we truncate it to the first 1000 sample pairs.
\item For synthesis agents, we generate datasets of 500 samples by prompting the model to produce 10 samples at a time and concatenating the results over 50 iterations. This mitigates potential quality degradation from long context windows.
\end{itemize*}
\input{tabs/main_result_ddr}

%% file: figs/tex/benchmark.tex
\begin{figure*}[t]
    \centering
    \includegraphics[width=1.0\linewidth]{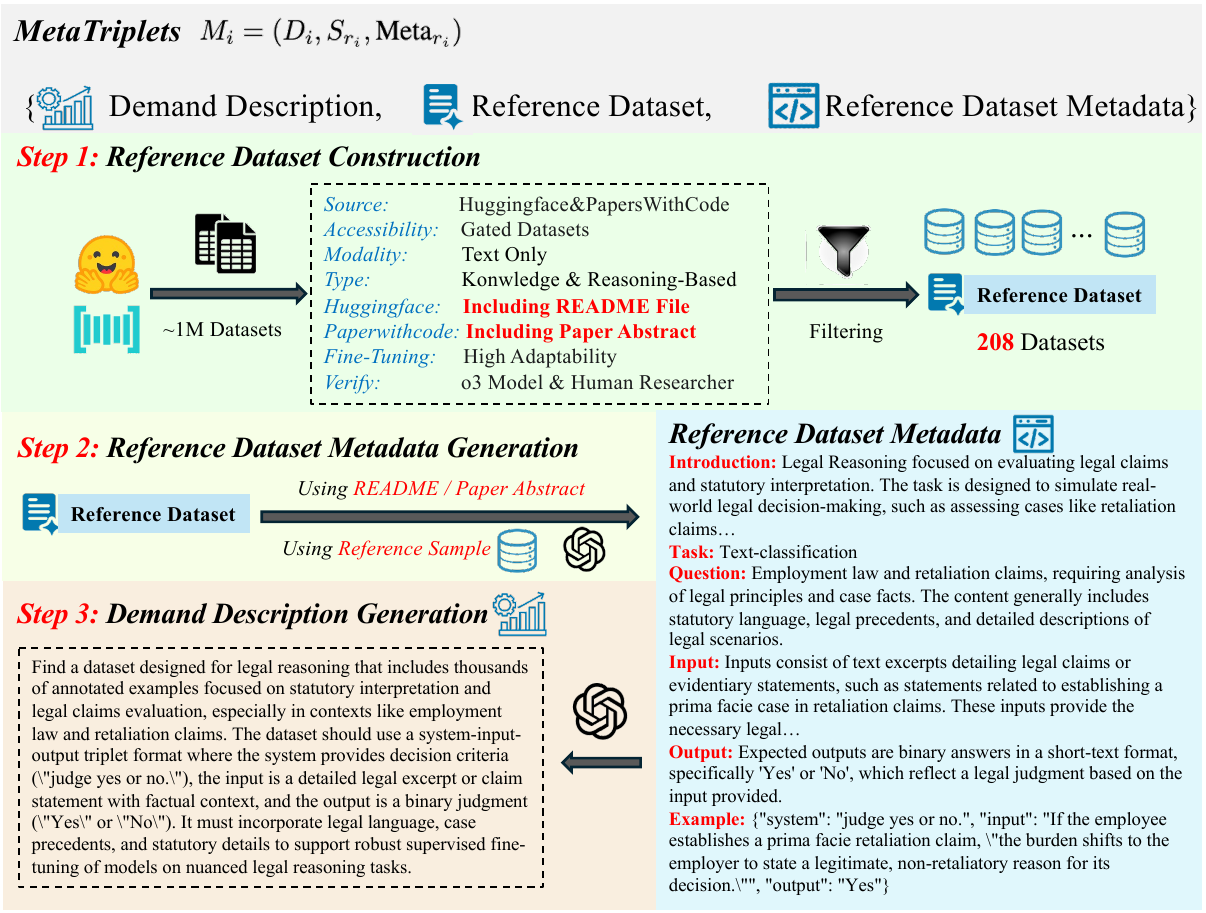}
    \caption{\textbf{Curation pipeline of the \ourbench benchmark.} From an initial dataset of over 1 million candidates, we first apply a series of filtering rules to curate a final reference set of 208 instances. We then utilize the state-of-the-art o3 model to process the associated README files and data samples, generating metadata across six distinct dimensions. Finally, the o3 model synthesizes this metadata to generate the corresponding dataset demands.}
    \label{fig:benchmark}
\end{figure*}

%% file: tabs/dataset_info.tex
\begin{table}[t]
\centering
\begin{tabular}{l c cc cc c} 
\toprule

\multirow{2}{*}{\textbf{Task}} & \multirow{2}{*}{\textbf{Metric}} & \multicolumn{2}{c}{\textbf{Num(Knowledge)}} & \multicolumn{2}{c}{\textbf{Num(Reasoning)}} & \multirow{2}{*}{\textbf{Num(pro)}} \\

& & \includegraphics[width=0.04\linewidth]{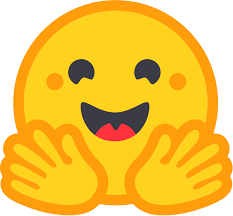} & \includegraphics[width=0.04\linewidth]{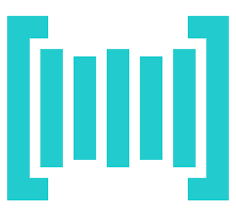} & \includegraphics[width=0.04\linewidth]{figs/images/hf.png} & \includegraphics[width=0.04\linewidth]{figs/images/pc.png} & \\

\midrule %
Multiple Choice & Accuracy & 9 & 4 & 4 & 10 & 5 \\
Text Generation & BLEU & 6 & 2 & 9 & 21 & 3 \\
Text Summarization & ROUGH & 2 & 1 & 0 & 8 & 3 \\
Question Answering & F1, Exact Match & 4 & 9 & 10 & 25 & 3 \\
Text Classification & Accuracy & 9 & 3 & 23 & 23 & 3 \\
Language Translation & BLEU & 1 & 1 & 14 & 10 & 3 \\

\bottomrule 
\end{tabular}
\caption{\ourbench comprises \ourbenchnum dataset demands derived from six categories of real-world NLP datasets from two distinct sources, evaluated using diverse metrics. The reference datasets in \ourbench are divided into knowledge-based and reasoning-based tasks. \ourbenchmini is a subset containing 20 more challenging examples.}
\label{tab:dataset_info}
\end{table}

%% file: figs/tex/evaluation.tex
\begin{figure*}[t]
    \centering
    \includegraphics[width=1.0\linewidth]{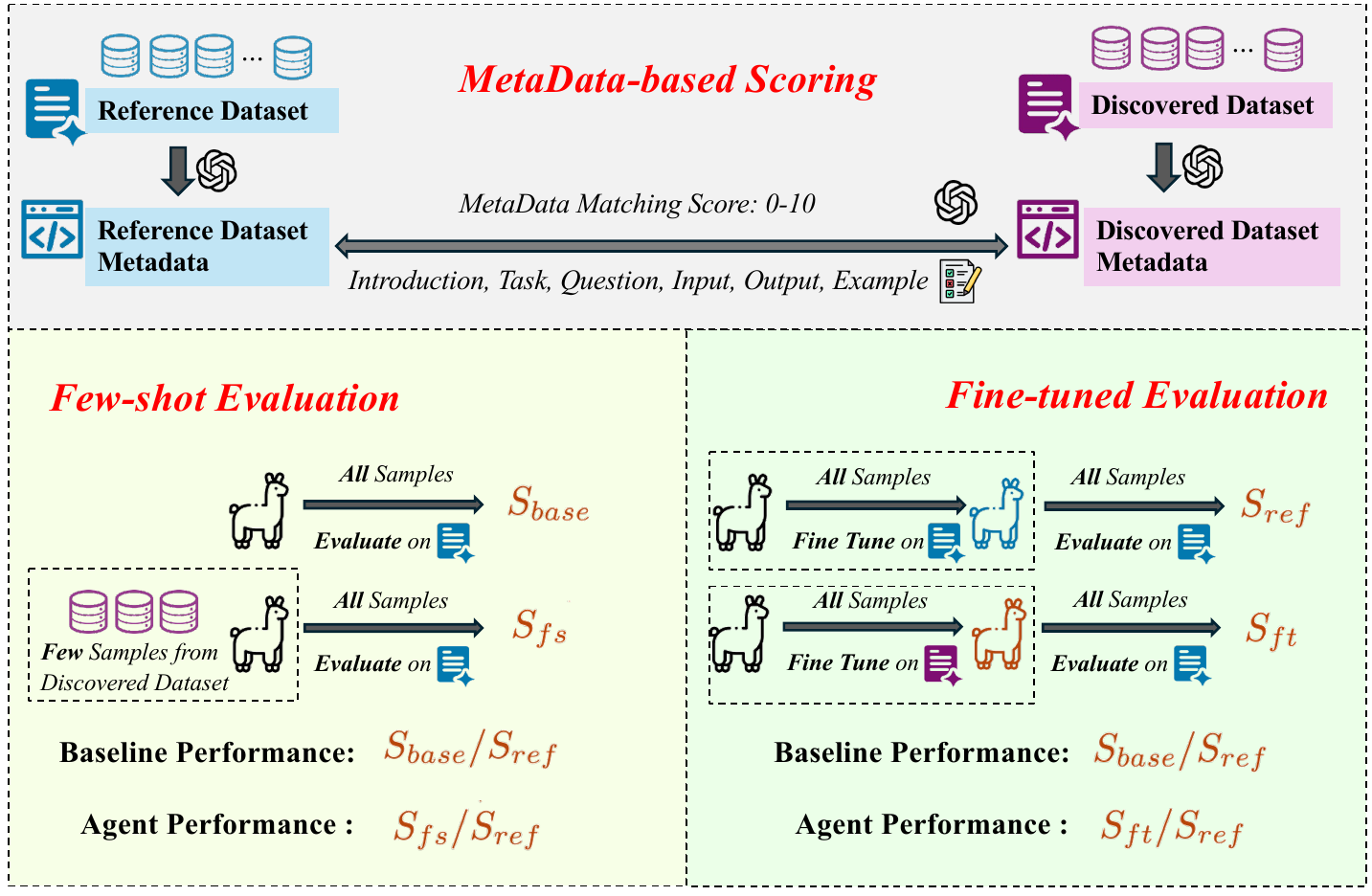}
    \caption{\textbf{The \ourbench evaluation methodology.} Our evaluation framework assesses data quality from three perspectives. We compute the average metadata similarity score between the discovered datasets' metadata (from \ourbaseline) and reference datasets' metadata across six dimensions using o3. For fine-tune performance, we measure the ratio of \llama model performance using discovered dataset exemplars ($S_{ft}$) to that using reference dataset exemplars ($S_{ref}$). This is compared against a baseline ($S_{base}/S_{ref}$), where $S_{base}$ is \llama zero-shot performance on reference dataset. The same evaluation is repeated with few-shot learning. All evaluations are performed using the \llama model. The specific metrics for each task are detailed in \Cref{tab:dataset_info}.}
    \label{fig:evaluation}
\end{figure*}

%% file: figs/tex/baseline.tex
\begin{figure*}[t]
    \centering
    \includegraphics[width=1.0\linewidth]{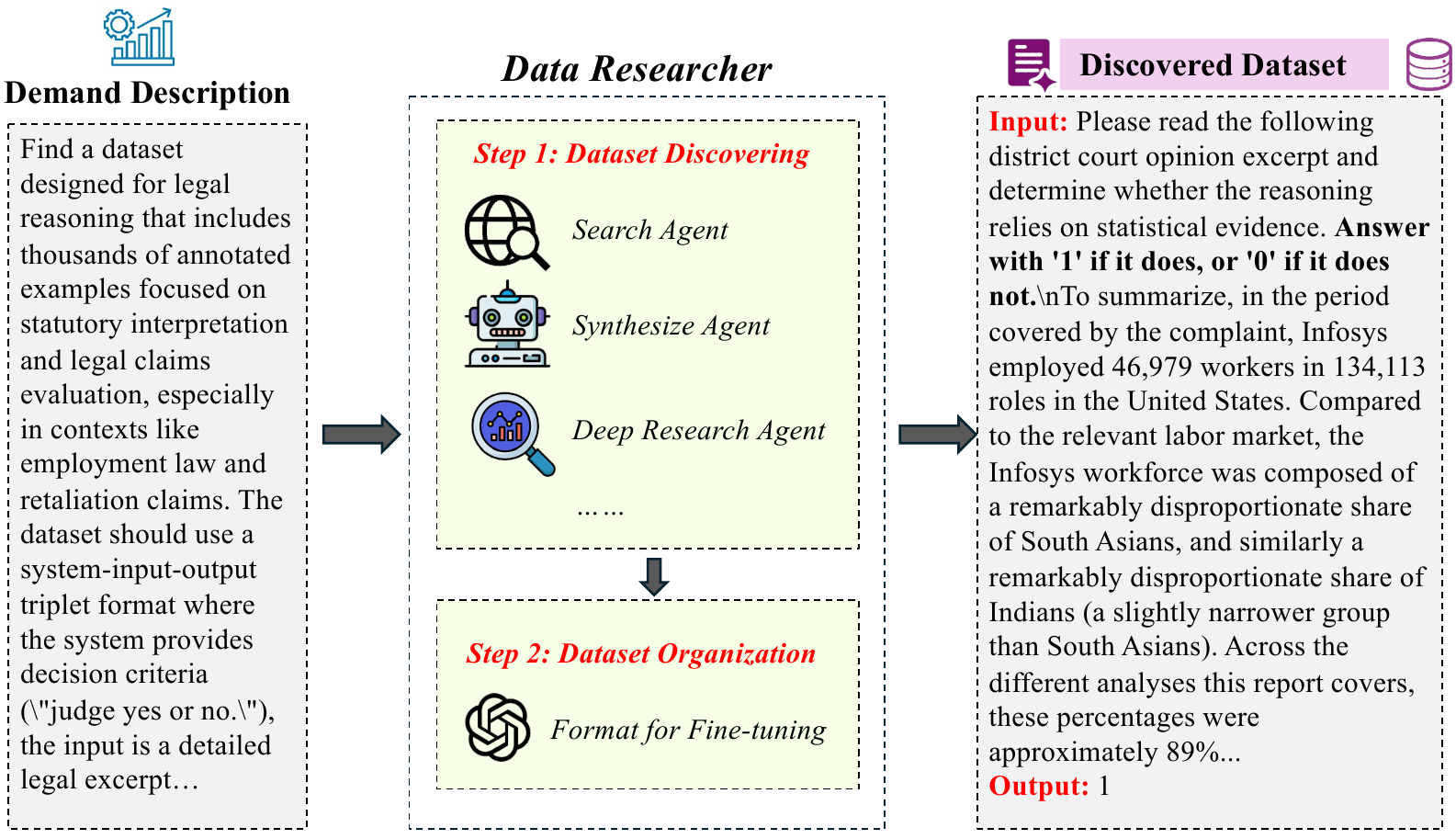}
    \caption{\textbf{Overview of the DataResearcher Baseline Workflow.} The process begins with dataset requirements extracted from a reference set, with explicit name information removed. The DataResearcher module takes these requirements and produces the discovered dataset through various means, such as searching for existing datasets or synthesizing a new one. Subsequently, OpenAI o3 generates corresponding metadata for discovered dataset, enabling a comparative analysis with the metadata of the reference dataset shown in \Cref{fig:evaluation}.}
    \label{fig:baseline}
\end{figure*}

%% file: tabs/main_result_ddr.tex
\begin{table}[t]
\centering
\resizebox{\textwidth}{!}{%
\begin{tabular}{llccccccccc}
\toprule

\multirow{2}{*}{\textbf{Agent}} & \multirow{2}{*}{\textbf{Method}} & \multicolumn{2}{c}{\textbf{DTP Evaluation}} & \multicolumn{7}{c}{\textbf{Metadata Evaluation}}  \\ \cline{3-11} 
 &  & \textit{Knowledge}(\%)$\uparrow$ & \textit{Reasoning}(\%)$\uparrow$ & Intro & Task & Ques & Input & Output & Example & Avg $\uparrow$  \\ 
 
\midrule

\textit{\textbf{Baseline}} & &  &  &  &  & & &  & &  \\ 

\multirow{1}{*}{} 
&  & 10.41 & 11.84 &  &  &  &  &  & &\\ 

\midrule
 
\textit{\textbf{Search-based}} & &  &  &  &  & & &  & &  \\ 

\multirow{4}{*}{GPT-4o-search} 
& \cellcolor{coolblue1}1 Shot & \cellcolor{coolblue1}9.82 & \cellcolor{coolblue1}7.25 & \multirow{4}{*}{5.7600}  & \multirow{4}{*}{6.6300} & \multirow{4}{*}{5.3600} &  \multirow{4}{*}{5.7700}& \multirow{4}{*}{5.3200} & \multirow{4}{*}{5.4100} & \multirow{4}{*}{5.7083} \\
& \cellcolor{coolblue2}3 Shots & \cellcolor{coolblue2}9.84 & \cellcolor{coolblue2}8.43 &  &  &  &  &  & &\\
& \cellcolor{coolblue3}5 Shots & \cellcolor{coolblue3}10.22 & \cellcolor{coolblue3}8.70 &  &  &  &  &  & &\\
& \cellcolor{coolblue4}Fine Tune & \cellcolor{coolblue4}\textbf{41.89} & \cellcolor{coolblue4}27.54 &  &  &  &  &  & &\\ 
 
\midrule
 
\multirow{4}{*}{GPT-4o-mini-search} 
& \cellcolor{coolblue1}1 Shot & \cellcolor{coolblue1}6.48 & \cellcolor{coolblue1}5.73 & \multirow{4}{*}{5.5000}  & \multirow{4}{*}{6.4400} & \multirow{4}{*}{5.2900} &  \multirow{4}{*}{5.7000}& \multirow{4}{*}{4.8800} & \multirow{4}{*}{5.3300} & \multirow{4}{*}{5.5233} \\
& \cellcolor{coolblue2}3 Shots & \cellcolor{coolblue2}6.45 & \cellcolor{coolblue2}7.89 &  &  &  &  &  & &\\
& \cellcolor{coolblue3}5 Shots & \cellcolor{coolblue3}10.38 & \cellcolor{coolblue3}8.39 &  &  &  &  &  & &\\
& \cellcolor{coolblue4}Fine Tune & \cellcolor{coolblue4}12.12 & \cellcolor{coolblue4}17.35 &  &  &  &  &  & &\\ 
 
\midrule

\textit{\textbf{Synthesis-based}} & &  &  &  &  &  &  & &&\\ 

\multirow{4}{*}{OpenAI o3 w/ ref} 
& \cellcolor{coolblue1}1 Shot & \cellcolor{coolblue1}10.25 & \cellcolor{coolblue1}17.38 & \multirow{4}{*}{8.6300}  & \multirow{4}{*}{8.7100} & \multirow{4}{*}{8.1100} &  \multirow{4}{*}{9.0100}& \multirow{4}{*}{9.3600} & \multirow{4}{*}{8.3200} & \multirow{4}{*}{\textbf{8.6900}} \\
& \cellcolor{coolblue2}3 Shots & \cellcolor{coolblue2}21.81 & \cellcolor{coolblue2}32.14 &  &  &  &  &  & &\\
& \cellcolor{coolblue3}5 Shots & \cellcolor{coolblue3}23.91 & \cellcolor{coolblue3}28.92 &  &  &  &  &  & &\\
& \cellcolor{coolblue4}Fine Tune & \cellcolor{coolblue4}\underline{38.98} & \cellcolor{coolblue4}\textbf{72.70} &  &  &  &  &  & &\\ 

\midrule

\multirow{4}{*}{OpenAI o3 w/o ref} 
& \cellcolor{coolblue1}1 Shot & \cellcolor{coolblue1}10.16 & \cellcolor{coolblue1}12.26 & \multirow{4}{*}{8.5800}  & \multirow{4}{*}{8.7400} & \multirow{4}{*}{8.1100} &  \multirow{4}{*}{8.8000}& \multirow{4}{*}{9.1700} & \multirow{4}{*}{8.0400} & \multirow{4}{*}{\underline{8.5730}} \\
& \cellcolor{coolblue2}3 Shots & \cellcolor{coolblue2}17.25 & \cellcolor{coolblue2}25.53 &  &  &  &  &  & &\\
& \cellcolor{coolblue3}5 Shots & \cellcolor{coolblue3}14.81 & \cellcolor{coolblue3}19.44 &  &  &  &  &  & &\\
& \cellcolor{coolblue4}Fine Tune & \cellcolor{coolblue4}37.94 & \cellcolor{coolblue4}\underline{67.25} &  &  &  &  &  & &\\ 
 
 \bottomrule
\end{tabular}%
}
\caption{\textbf{Main results on the \ourbench benchmark.} We report normalized few-shot and fine-tuning scores on Knowledge and Reasoning tasks using DTP (Downstream Task Evaluation), alongside metadata similarity scores. Synthesis-based agents (OpenAI o3) demonstrate superior performance on reasoning tasks, whereas search-based agents (GPT-4o-search) excel at knowledge-based tasks. Display Best results are displayed in \textbf{bold}, the second-best results with an \underline{underline}.}
\label{tab:main_result_ddr}
\end{table}

%% file: sec/experiment.tex
\section{Experiments}

Our experiments are designed to rigorously assess the capabilities of \ourbaseline in demand-driven dataset discovery. We present results on both the comprehensive \ourbench benchmark and its challenging \ourbenchmini subset, revealing critical insights into the strengths and weaknesses of current search-based, synthesis-based and deep research approaches.
\subsection{Experimental Setup}
We evaluate leading agents on our \ourbench with 208 tasks and \ourbenchmini with 20 tasks benchmarks. The agents include search-based \ourbaseline like GPT-4o-search, synthesis-based \ourbaseline like OpenAI o3 with and without reference examples and deep research based \ourbaseline like OpenAI DeepResearch. Evaluation is twofold: (1) Metadata-based Evaluation that reflects dataset discovery instruction-following capabilities, and (2) Downstream Task Performance (DTP) that reflects the overall data performance of the discovered set. In our experiments, we calculate six distinct metrics tailored to different task categories:
\begin{itemize*}
    \item \textbf{Accuracy} measures the proportion of correct predictions, reflecting models’ ability to identify the correct option or label.
    \item \textbf{F1-Score}~\cite{joshi2017triviaqa} computes the harmonic mean of precision and recall, providing a balanced assessment of model performance, especially when dealing with partial matches or token-level evaluation.
    \item \textbf{Exact Match}~\cite{joshi2017triviaqa} evaluates the percentage of predictions that exactly match the ground-truth answers, offering a strict assessment criterion.
    \item \textbf{BLEU}~\cite{papineni2002bleu} measures n-gram overlap between generated and reference text, assessing the quality and fluency of generated content.
    \item \textbf{SacreBLEU}~\cite{keenan2017sacre} provides a standardized and reproducible version of BLEU scoring, ensuring consistent evaluation.
    \item \textbf{ROUGE}~\cite{lin2004rouge} calculates recall-oriented overlap of n-grams and longest common subsequences, specifically designed for evaluating text summarization quality and content preservation.
\end{itemize*}

\subsection{Main Results and Analysis}
\paragraph{Performance on the \ourbench Benchmark} As shown in \Cref{tab:main_result_ddr}, agent performance on \ourbench highlights a clear dichotomy in capabilities based on task-cognitive demands. For knowledge-based demands, search-based \ourbaseline demonstrate significant advantages where GPT-4o-search agent achieves the highest fine-tuning score of 42\%. Conversely, for reasoning-based tasks synthesis-based agents are undoubtedly superior, with the OpenAI o3 w/ ref agent attaining the highest fine-tuning score of 73\%. Notably, we observe that few-shot evaluation results exhibit outcomes that are closely aligned with fine-tuning experiments across both task categories, which suggests a practical implication: since fine-tuning experiments are computationally expensive and time-consuming, few-shot evaluation can serve as an efficient preliminary assessment method to rapidly detect and compare \ourbaseline capabilities before committing to full fine-tuning procedures.

In metadata evaluation, we reveal that synthesis-based methods significantly outperform across output metrics, which reveals the core advantage of synthesis-based methods in fine-tuning tasks: their ability to generate more aligned output data, thereby providing models with sufficient learning material to master reasoning pathways from input to output. Furthermore, we identify that the primary factor limiting search-based methods' performance lies in the fact that retrieved existing datasets often cannot align with the instruction as precisely as synthesized data, resulting in relatively weaker data research instruction-following capabilities.

\begin{tcolorbox}[
colback=gred!25,
colframe=gred!95,
]
\textbf{Takeaway:} Our evaluation reveals a task-dependent performance dichotomy where search-based agents excel at knowledge-intensive tasks while synthesis-based agents demonstrate superior performance on reasoning-heavy tasks, with the latter's advantage stemming from their ability to generate reasoning-rich, more instruction-aligned output data.
\end{tcolorbox}

\paragraph{Analysis of Learning Paradigms} Our evaluation framework also conducts detailed analysis of different learning strategies. We demonstrate that 1-shot, 3-shot, and 5-shot results maintain consistent relative trends with fine-tuning, although some 5-shot experiments fail to follow expected shot-scaling patterns. This is primarily attributed to the small-scale models we employ being unable to maintain effective long-range attention mechanisms when processing longer context windows~\cite{schaeffer2023emergent}. Notably, among few-shot settings, 3-shot demonstrates the most stable and representative performance, achieving a favorable balance between computational efficiency and performance outcomes.


\paragraph{Performance on the \ourbenchmini Subset} To explore the performance boundaries of state-of-the-art deep research agents, we conduct experiments on the \ourbenchmini subset. Results shown in \Cref{fig:bar} and \Cref{fig:radar} demonstrate that advanced deep research systems such as OpenAI DeepResearch, achieving a score of 0.2218, significantly outperform standard search methods like GPT-4o-mini-search, yet their overall performance remains relatively modest. Meanwhile, we observe that synthesis-based methods also exhibit significant performance degradation on \ourbenchmini, achieving only scores approaching 0.5, indicating that tasks challenging for search-based approaches similarly pose substantial difficulties for synthesis methods, which we discuss in detail in \Cref{sec:corner_case}.

\begin{tcolorbox}[
colback=gred!25,
colframe=gred!95,
]
\textbf{Takeaway:} On the difficult \ourbenchmini subset, advanced deep research systems, while superior to standard search agents, still perform poorly with a top score around 0.22. The substantial performance degradation observed across both approaches highlights the existence of fundamentally difficult data discovery tasks that challenge current \ourbaseline regardless of their methodology.
\end{tcolorbox}

\input{figs/tex/bar}

%% file: figs/tex/bar.tex
\begin{figure*}[t]
    \centering
    \includegraphics[width=1.0\linewidth]{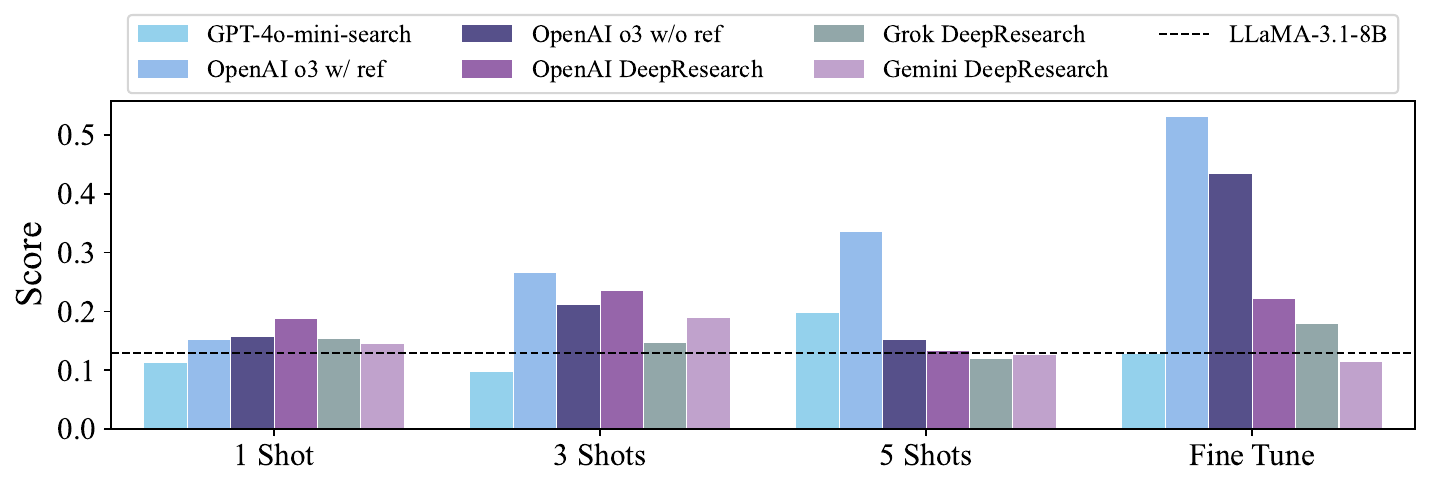}
    \caption{\textbf{Performance comparison of various agent systems on the \ourbenchmini subset across few-shot and fine-tuning settings.} Synthesis agents based on o3 perform exceptionally well, and the DeepResearch system generally outperforms the GPT-4o search system. The dashed line indicates the performance of \llama baseline on test set.}
    \label{fig:bar}
\end{figure*}

%% file: sec/analysis.tex
\input{figs/tex/case1}
\input{figs/tex/case2}
\input{figs/tex/case3}

\section{Analysis}

This section delves into specific cases to provide a qualitative understanding of the performance patterns observed in our experiments. By examining individual examples from \Cref{fig:case1}, \Cref{fig:case2} and \Cref{fig:case3}, we highlight the distinct behaviors of search, synthesis, and deep research methodologies, particularly focusing on the fine-tuning performance.

\subsection{Knowledge vs. Reasoning}
\Cref{fig:case1} offers a clear contrast between how different methods handle knowledge and reasoning-based tasks \cite{ravichander2019question}. For reasoning-centric requirements, synthesis-based methods construct richly detailed data with explicit thought processes, which guides the fine-tuned model toward more logical analysis. In contrast, for knowledge-based tasks, search methods excel by retrieving diversified information with broader coverage and higher knowledge density. This enables the fine-tuned model to access richer knowledge sources, thereby exhibiting stronger knowledge coverage and response accuracy on the reference set.

This reveals a key behavioral pattern: the breadth advantage of search methods allows them to capture long-tail knowledge distributions, making models more robust for diverse, fine-grained factual queries. Conversely, the depth advantage of synthesis methods lies in the ability of powerful LLMs to construct highly structured, logically coherent reasoning data, making it ideal for training models on tasks that depend on generalizable logic rather than pure fact retrieval.

\subsection{Deep Research Methodology}
As observed in \Cref{fig:case2}, the deep research methodology, with its iterative information gathering and reasoning-guided exploration, retrieves data of significantly higher quality than single-shot search or generation. Under identical query conditions\cite{dao2023vnhsgevietnamesehighschool}, the resulting dataset format demonstrates superior alignment with the reference set's characteristics, exhibiting greater comprehensiveness and analytical depth. This core advantage stems from its multi-round process, which constructs a more refined analytical perspective and uncovers high-quality datasets that are otherwise difficult to find. This enhances not only the breadth of information but, more importantly, the logical coherence of the resulting dataset, leading to clear performance gains in the final fine-tuning evaluation.

\subsection{Limitations of Current Methods for Corner Cases}
\label{sec:corner_case}
Despite their power, existing search and synthesis methods are fundamentally constrained by the data distributions they were trained on. \Cref{fig:case3} illustrates a ``corner case''—a task scenario so niche that constructing a suitable training dataset from existing sources is nearly impossible without directly plagiarizing the reference set~\cite{hoham2024medconceptsqa}. Consequently, the resulting fine-tuning performance is markedly poor.

This limitation is rooted in the data-dependent nature of current agents. Search methods are limited to what is indexed, and synthesis methods are limited to patterns seen during training. Because real-world data distributions are imbalanced and often underrepresent the niche, ``corner case'' scenarios, models lack the necessary prior knowledge to perform well. This demonstrates an inherent limitation in agents that rely solely on existing data distributions and calls for the development of more flexible and adaptive solutions.

\begin{tcolorbox}[
colback=gred!25,
colframe=gred!95,
]
\textbf{Takeaway:} Case studies of fine-tuning results confirm that search agents are superior for knowledge-based tasks due to their breadth, while synthesis agents excel at reasoning-based tasks by creating structured, logical examples. Advanced Deep Research methods outperform simple search by using iterative, reasoning-guided exploration. However, all current agent methodologies fail on ``corner cases'' that fall outside of existing data distributions, revealing a critical limitation.
\end{tcolorbox}

%% file: figs/tex/case1.tex
\begin{figure*}[!t]
    \centering
    \includegraphics[width=1.0\linewidth]{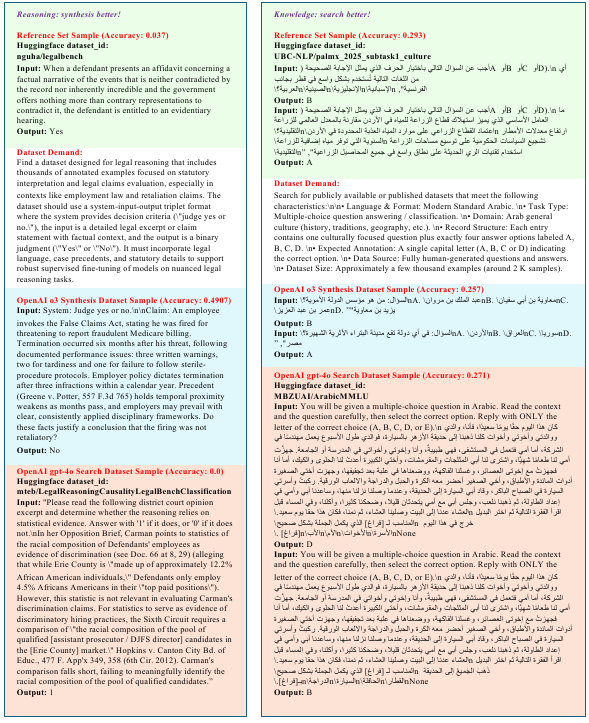}
    \caption{\textbf{Case study comparing data construction for reasoning-based versus knowledge-based tasks.} The left example (a legal reasoning task) shows that a synthesis agent generates a structured, high-quality output. The right example (an Arabic cultural classification task) demonstrates that a search agent successfully finds a highly relevant existing dataset with broad factual coverage.}
    \label{fig:case1}
\end{figure*}

%% file: figs/tex/case2.tex
\begin{figure*}[!t]
    \centering
    \includegraphics[width=0.9\linewidth]{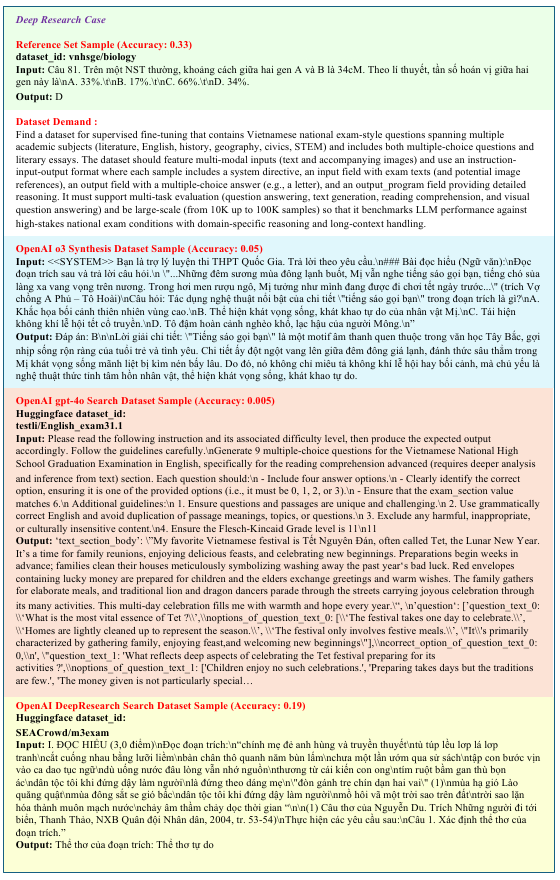}
    \caption{In this case, deep research agent demonstrates superior performance over \ourbaseline with search and synthesis agents.}
    \label{fig:case2}
\end{figure*}


%% file: figs/tex/case3.tex
\begin{figure*}[!t]
    \centering
    \includegraphics[width=1.0\linewidth]{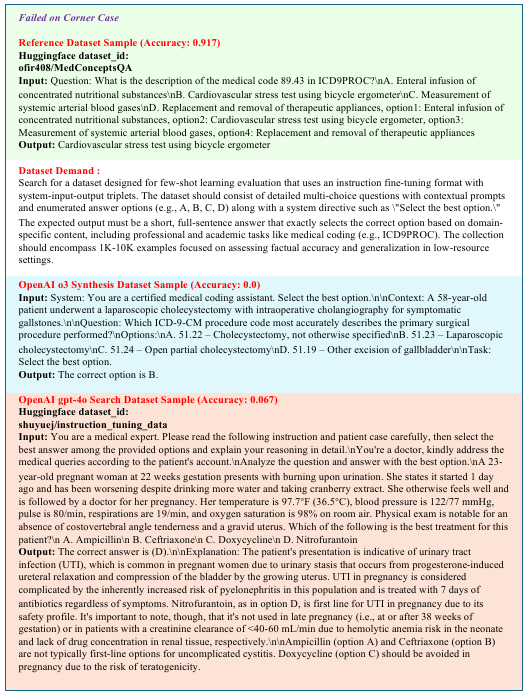}
    \caption{Performance comparison in a challenging corner case. All evaluated methods show degraded performance in this scenario.}
    \label{fig:case3}
\end{figure*}


%% file: sec/conclusion.tex
\section{Discussion}

\subsection{Conclusion}

In this work, we introduce \ourbench, the first comprehensive benchmark designed to systematically evaluate the capabilities of agentic systems in demand-driven dataset discovery and synthesis. By curating 208 tasks from real-world platforms and classifying them into knowledge-based and reasoning-based categories, we create a robust framework for assessing agent performance. Our multi-faceted evaluation protocol, which combines metadata alignment with downstream task performance via few-shot learning and fine-tuning, provides a nuanced view of current system capabilities.

Our extensive experiments reveal a critical performance gap in state-of-the-art agents. We discovere a clear specialization: search-based agents excel at knowledge-intensive tasks, while synthesis-based agents dominate complex reasoning challenges. Notably, even advanced deep research systems achieve a top fine-tuning score of only around 22\% on our difficult \ourbenchmini subset, whereas synthesis agents demonstrate surprising robustness, more than doubling this performance. These findings underscore that while promising, current agentic systems have significant room for improvement. Our benchmark and findings lay a crucial foundation for guiding the development of the next generation of AI-powered research assistants.

\subsection{Limitation and Future Work}
While DeepDatasetResearch provides a foundational framework, we acknowledge its limitations and see several exciting avenues for future research.

\paragraph{Automating Web-Scale Data Curation} Our current benchmark relies on datasets from structured repositories like HuggingFace and PaperswithCode. A key next step is to develop agents capable of automatically discovering, parsing, and formatting data from the unstructured web—including academic papers, code repositories, and blogs. This would test an agent's ability to handle noisy, diverse, and unstructured information, moving closer to a true automated research workflow.

\paragraph{Exploring Open-Source Model Capabilities for Data Synthesis} Our current synthesis agent evaluation primarily relies on state-of-the-art closed-source reasoning models OpenAI o3, which ensures high-quality synthesized data but incurs relatively high computational costs and API expenses. An important future research direction involves systematically evaluating the performance of open-source large language models in data synthesis tasks, including but not limited to LLaMA, Mistral, and other model families. Additionally, developing agent systems based on open-source models for data synthesis would help lower the cost barriers of data construction, enabling broader access to automated data synthesis capabilities for researchers and institutions. This exploration would not only democratize data construction capabilities but also provide crucial insights into the strengths and limitations of different model architectures in data synthesis tasks.

\paragraph{Hybrid Agents Integrating Search and Synthesis} Our findings reveal a clear dichotomy between search and synthesis agents. A natural evolution would be the development of hybrid agents as the \ourbaseline that intelligently combine multiple strategies. Such \ourbaseline could first search for existing datasets, then apply reject sampling, data filtering, or quality enhancement methods to refine and synthesize the collected data. When existing datasets partially match demands, the \ourbaseline could supplement missing components through synthesis agents; when search fails, the \ourbaseline could collect relevant corpora and organize, annotate, and format these raw materials into structured datasets suitable for specific training tasks. This multi-stage integration approach promises a better balance between data quality and coverage.









%% file: sec/appendix.tex

\appendix
\section{Metadata}
\label{sec:metadata}
Our benchmark provides comprehensive metadata annotations for each task instance to facilitate systematic evaluation and analysis. Each MetaTriplet contains the following structured metadata components:
\begin{itemize}
    \item \textbf{Introduction}: The task and area description of this task instance.
    \item \textbf{Task}: The classification of task type.
    \item \textbf{Question}: Question Content Type - Describes the primary knowledge domains and content types covered by the questions in the test dataset.
    \item \textbf{Input}: Structured retrieval results and contextual information - Input consists of formatted search results containing metadata fields such as descriptions, display URLs, titles, actual URLs, and ranking information, along with potential tabular data, document snippets, and conversational dialogue history for multi-turn scenarios.
    \item \textbf{Output}: Direct factual answer format - Outputs are concise, definitive answers that directly address the question based on the provided context, formatted as complete statements such as 'The answer is [specific fact]' for factual queries, numerical values for arithmetic problems, and explicit acknowledgment when questions cannot be answered.
    \item \textbf{Example}: dataset instance example.
\end{itemize}

\section{Metadata Evaluation of \ourbenchmini}
We present the metadata results of six discovery approaches on \ourbenchmini through a radar chart, as shown in \Cref{fig:radar}. The chart clearly
  demonstrates significant performance differences between the two major method categories, particularly highlighting the comprehensive advantages of
  the Deep Research method in the Search Group across all dimensions, with especially excellent performance in Task and Output Alignment. The Deep
  Research approach brings comprehensive benefits, particularly in task and output aspects. Notably, generative methods perform most prominently in
  the Output and Intro dimensions, reflecting the inherent advantages of synthetic data in output quality and instruction-following accuracy for
  tasks. Furthermore, the comparison between OpenAI o3 w/ ref and w/o ref validates the positive impact of reference samples on improving data
  generation quality.
\input{figs/tex/radar}
\section{Prompts}

\subsection{Prompts for Dataset Curation}
This prompt instructs OpenAI o3 to generate a comprehensive dataset demand description by transforming extracted metadata into generic, discoverable descriptions that omit specific identifiers while preserving all essential characteristics needed for effective dataset retrieval.
\begin{tcolorbox}[colframe=black, title={Prompt for demand description generation}]
Imagine you are performing a dataset search task, and the ultimate ground truth (the most ideal dataset found) is the dataset I provided to you. Now, based on your extraction of this dataset's metadata in the previous conversation, generate a prompt for the dataset search task (this prompt will be passed to a powerful deep research agent to complete the task). This prompt needs to include all the metadata from the previous conversation, except for the example.\\\\
Previous metadata extraction:
\{metadata\_json\}\\

Please generate a search prompt that a research agent could use to find this exact type of dataset. The prompt should be comprehensive and include all the key characteristics that would help identify this specific dataset type.\\\\
IMPORTANT REQUIREMENTS:\\
1. DO NOT mention the original dataset name or any specific dataset identifiers\\
2. Include all metadata information EXCEPT the example field\\
3. For dataset size/samples count, use approximate ranges rather than exact numbers (e.g., "around 10K samples" instead of exact counts)\\
4. Focus on the task type, domain, input/output characteristics, and source information\\
5. Make the prompt generic enough that it could find similar datasets, not just the specific one\\\\
Output the search prompt directly without any additional formatting or explanation.
\end{tcolorbox}

\subsection{Prompts for Data Researcher}
This prompt instructs AI to search Hugging Face for publicly accessible datasets and return a specified number of dataset IDs in strict JSON format.
\begin{tcolorbox}[colframe=black, title={Prompt for search agent}]
\{agent\_query\} \\\\
IMPORTANT: You must search for publicly accessible datasets from Hugging Face and return exactly \{NUM\_SEARCH\} suitable dataset IDs. \\\\
SEARCH STRATEGY:\\
- First, search for datasets that closely match the query\\
- If you cannot find enough datasets, gradually relax the search criteria to find more potential datasets\\
- Include both popular and less popular datasets that might be relevant\\
- Consider datasets with similar tasks, domains, or data types\\\\
OUTPUT FORMAT REQUIREMENTS:\\
You MUST output your response in EXACTLY this JSON format - do not include any other text or explanations:\\\\
\{'search\_datasets': ['dataset\_id\_1', 'dataset\_id\_2', 'dataset\_id\_3', 'dataset\_id\_4', 'dataset\_id\_5', 'dataset\_id\_6', 'dataset\_id\_7', 'dataset\_id\_8', 'dataset\_id\_9', 'dataset\_id\_10']\} \\\\
- Use ONLY the exact format above\\
- Replace dataset\_id\_1, dataset\_id\_2, etc. with actual Hugging Face dataset IDs\\
- Ensure you provide exactly \{NUM\_SEARCH\} dataset IDs\\
- Do not include any text before or after the JSON\\
- The JSON must be valid and parseable

\end{tcolorbox}

This fine-tuning data extraction prompt guides AI to analyze dataset structure and establish conversion rules for transforming it into fine-tuning format with input-output fields, requiring a mandatory task-specific instruction template.
\begin{tcolorbox}[colframe=black, title={Prompt for fine tuning data extraction from huggingface datasets}]
You are tasked with analyzing a dataset and determining how to convert it into a fine-tuning format for language models. The fine-tuning data should have exactly two fields: "input" and "output".\\\\
\{readme\_section\}\\
\{config\_section\}\\
\{split\_section\}\{sample\_section\}\\\\
Based on the README content (if available) and the sample data, please analyze and provide the following information:\\\\
1. **Selected Config**: Which config/subset was selected from the dataset (output "None" if there's only one config)\\
2. **Selected Split**: Which split was selected (train/test/validation/etc.)\\
3. **Conversion Rules**: Describe the rules for converting the dataset samples into input/output format for fine-tuning, including:\\
   - Which columns/fields should be combined for the "input"\\
   - Which columns/fields should be used for the "output" \\ 
   - **MUST include appropriate instruction text** (e.g., "Please answer the following question:", "Classify the sentiment:", "Translate the text:", etc.)\\
   - The exact format and order for combining the fields\\
   - Any preprocessing needed for the data\\\\

**IMPORTANT REQUIREMENTS:**\\
- The "instruction\_template" field MUST contain a clear, specific instruction appropriate for the task\\
- The instruction should tell the model exactly what to do (e.g., answer questions, classify, translate, summarize, etc.)\\
- Do NOT use "null" for instruction\_template - always provide a meaningful instruction\\
- The input should clearly guide the model on what output is expected\\\\
Please provide your response in the following JSON format:\\\\
\{\\
    "selected\_config": "config\_name or None",\\
    "selected\_split": "split\_name",\\
    "conversion\_rules": \\\{\\
        "input\_components": ["list of field names to include in input (use exact field names from sample)"],\\
        "output\_components": ["list of field names to include in output (use exact field names from sample)"],\\
        "instruction\_template": "REQUIRED: Clear instruction text telling the model what to do",\\
        "input\_format": "detailed description of how to format the input",\\
        "output\_format": "detailed description of how to format the output",\\
        "example\_conversion": \\\{\\
            "input": "example input with instruction based on the sample",\\
            "output": "example output based on the sample"\\
        \}\\
    \}\\
\}\\\\
Important: Make sure the conversion makes sense for language model fine-tuning and follows common patterns for instruction-following datasets. The instruction\_template is MANDATORY and should be task-specific.

\end{tcolorbox}

This synthesis agent prompt tasks AI with directly synthesizing a specified number of high-quality training samples based on data requirements, aiming to create superior fine-tuning data compared to existing searched datasets.
\begin{tcolorbox}[colframe=black, title={Prompt for synthesis agent}]
You are a specialized expert in fine-tuning data synthesis. You have the following dataset search requirement: \{agent\_query\} \\\\
Your task is to directly synthesize \{num\_data\} corresponding examples based on this requirement. The goal is to create synthetic data that, when used for fine-tuning a large language model, will achieve better performance than fine-tuning on existing datasets found through the search.\\\\
Here is a reference example for guidance: \{example\_data\} \\\\
You MUST output exactly \{num\_data\} samples in JSON list format, where each sample contains only "input" and "output" fields, following this exact format:\\

  \{
    "input": "...",
    "output": "..."
  \},
  \{
    "input": "...",
    "output": "..."
  \},
  ...\\
  
Important requirements:\\
1. Generate EXACTLY \{num\_data\} examples\\
2. Each example must have only "input" and "output" fields\\
3. Follow the task type and domain specified in the search requirement\\
4. Use the reference example to understand the expected format and style\\
5. Ensure diversity across your generated examples\\
6. Focus on creating high-quality data that will improve model performance through fine-tuning

\end{tcolorbox}

\subsection{Prompts for Metadata Evaluation}

This prompt instructs AI to analyze a HuggingFace dataset's README and sample data to extract comprehensive metadata including task type, domain, input/output descriptions, and source information in structured JSON format.
\begin{tcolorbox}[colframe=black, title={Prompt for dataset metadata generation}]
You will see a README file introduction from a HuggingFace dataset and one sample from it. You need to output the following content based on these materials. Please output in JSON format.\\\\
README content:\\\{readme\_content\}\{sample\_section\}\\\\
Please analyze the dataset based on both the README and the sample (prioritize the sample when there are conflicts, as the README might be vague while we've ensured all samples in the dataset are similar to the provided one), and output the following metadata in JSON format:\\\\
\{\\
    "introduction": "A one-sentence introduction of the dataset content, concise and clear, including key information about task, domain, input and output",\\
    "task": "Directly output one of: multiple-choice, question-answering, summarization, text-classification, text-generation, translation",\\
    "domain": "Directly output the domain the dataset content involves, such as: aerospace, finance, linguistics, politics, sociology, biology, etc.",\\
    "input": "Directly output the dataset's input content, including its language, such as: an English news text for translation, a multiple-choice question in French philosophy domain, etc. (Consider both sample and README, don't be limited by single sample's domain, but also don't be too broad like README)",\\
    "output": "Directly output the dataset's output content, including its language, such as: a number 0 or 1, a letter A/B/C/D, translated Italian text, etc.",\\
    "source": "Directly output the dataset's source: real-world, human-generated, machine-generated, etc.",\\
    "example": "Directly extract the sample provided in the prompt and put it here",\\
    "samples\_count": \{samples\_count\}
\\\}\\

Important: Please strictly follow the above JSON format and provide a comprehensive analysis based on both README and sample data.

\end{tcolorbox}

This metadata evaluation prompt tasks AI with comparing two dataset metadata objects across multiple dimensions and providing numerical similarity scores (0-10) for each dimension along with an overall average score in JSON format.

\begin{tcolorbox}[colframe=black, title={Prompt for metadata evaluation}]
I need you to compare two dataset metadata and score their matching degree across the following dimensions.\\\\
Dimension descriptions:\\
- introduction: Dataset introduction and overview\\
- task: Task type (e.g., text-classification, question-answering, etc.)\\
- domain: Domain field (e.g., finance, politics, biology, etc.)\\
- input: Description of input content\\
- output: Description of output content\\
- source: Data source (e.g., human-generated, machine-generated, etc.)\\
- example: Sample data\\
- samples\_count: Number of samples\\\\

Original dataset metadata:\\
\{original\_metadata\}\\

Search dataset metadata:\\
\{search\_metadata\}\\\\
Please score each dimension on a scale of 0-10 for matching degree, where:\\
- 10 points: Complete match or highly similar\\
- 0 points: Complete mismatch or opposite\\
- Output an integer score. If a dimension is missing or meaningless in one or both metadata, mark it as null\\

Please output the result strictly in the following JSON format:\\\\
\{\\
    "introduction": score or null,\\
    "task": score or null,\\
    "domain": score or null,\\
    "input": score or null,\\
    "output": score or null,\\
    "source": score or null,\\
    "example": score or null,\\
    "samples\_count": score or null,\\
    "average": average score (excluding null values) or null\\
\}\\\\
Note:\\
1. Only output JSON format, do not include any other text\\
2. Scores must be numbers between 0-10 or null\\
3. average is the mean of all non-null scores

\end{tcolorbox}
\section{Config for Supervised Fine-tuning}
Below is the standard configuration file for supervised fine-tuning using the LlamaFactory framework, based on the Llama-3.1-8B model with bfloat16 precision and full-parameter fine-tuning strategy:
\begin{tcolorbox}[colframe=black, title={Config for supervised fine-tuning}]
bf16: true\\
cutoff\_len: 4096\\
dataloader\_num\_workers: 0\\
dataset: \{dataset\_id\}\\
ddp\_timeout: 180000000\\
deepspeed: examples/deepspeed/ds\_z3\_config.json\\
do\_train: true\\
finetuning\_type: full\\
gradient\_accumulation\_steps: 2\\
learning\_rate: 1.0e-05\\
logging\_steps: 10\\
lr\_scheduler\_type: cosine\\
max\_samples: 1000\\
model\_name\_or\_path: models/LLama3/Llama-3.1-8B\\
num\_train\_epochs: 3.0\\
output\_dir: LLaMA-Factory/results/\{task\_id\}/saves\\
overwrite\_cache: true\\
overwrite\_output\_dir: true\\
per\_device\_train\_batch\_size: 1\\
plot\_loss: true\\
preprocessing\_num\_workers: 16\\
report\_to: none\\
resume\_from\_checkpoint: null\\
save\_only\_model: false\\
save\_steps: 1000\\
stage: sft\\
template: llama3\\
trust\_remote\_code: true\\
warmup\_ratio: 0.1
\end{tcolorbox}

%% file: figs/tex/radar.tex
\begin{figure*}[hbt]
    \centering
    \includegraphics[width=1.0\linewidth]{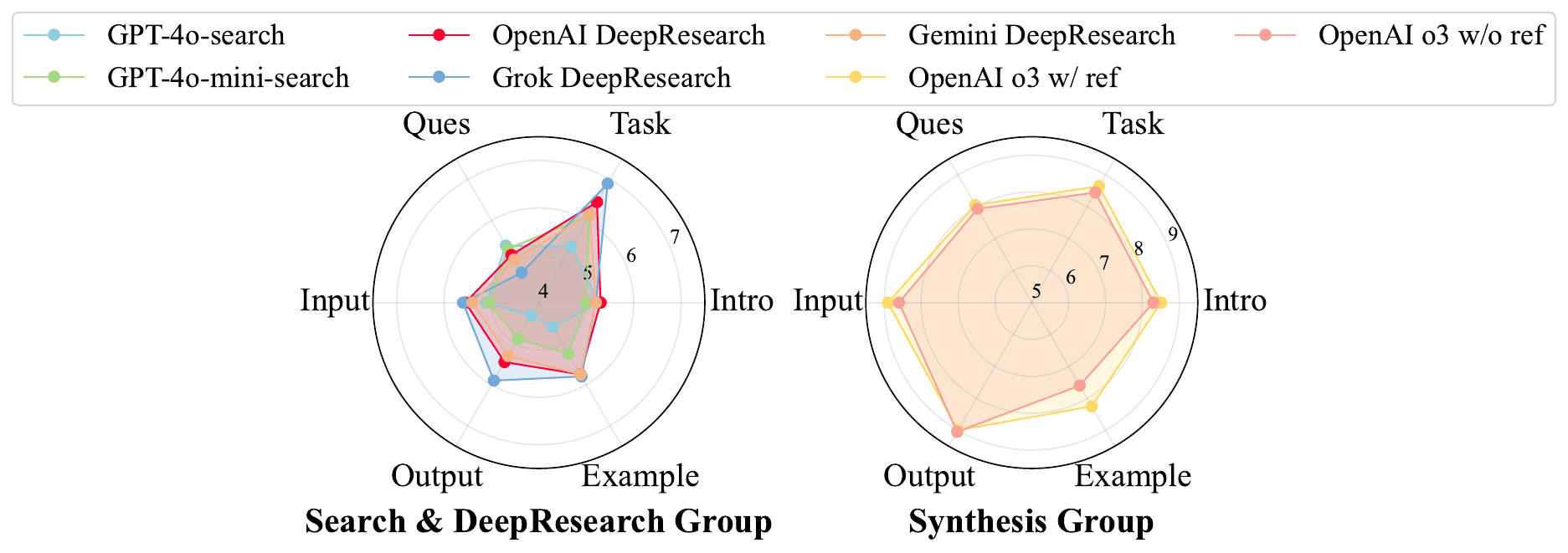}
    \caption{\textbf{Radar chart illustrating the average metadata alignment scores for the Search Group, Generation Group, and DeepResearch Group on the \ourbenchmini subset.} The Generation and DeepResearch groups achieve higher and more balanced scores, indicating superior semantic alignment with the ground-truth dataset requirements.}
    \label{fig:radar}
\end{figure*}